# Agent-Based Output Drift Detection for Breast Cancer Response Prediction in a Multisite Clinical Decision Support System

Xavier Rafael-Palou[1*], Jose Munuera[1], Ana Jimenez-Pastor[1], Richard Osuala[2], Karim Lekadir[2,3], Oliver Díaz[2,4]

[1] Research and Frontiers AI, Quibim, Valencia, Spain
[2] Departament de Matemàtiques i Informàtica, Universitat de Barcelona, Barcelona, Spain
[3] Institució Catalana de Recerca i Estudis Avançats (ICREA), Barcelona, Spain
[4] Computer Vision Center, Bellaterra, Spain
xavierrafael@quibim.com

**Abstract.** Modern clinical decision support systems can concurrently serve multiple, independent medical imaging institutions, but their predictive performance may degrade across sites due to variations in patient populations, imaging hardware, and acquisition protocols. Continuous surveillance of predictive model outputs offers a safe and reliable approach for identifying such distributional shifts without ground truth labels. However, most existing methods rely on centralized monitoring of aggregated predictions, overlooking site-specific drift dynamics.
We propose an agent-based framework for detecting drift and assessing its severity in multisite clinical AI systems. To evaluate its effectiveness, we simulate a multi-center environment for output-based drift detection, assigning each site a drift monitoring agent that performs batch-wise comparisons of model outputs against a reference distribution. We analyse several multi-center monitoring schemes, that differ in how the reference is obtained (site-specific, global, production-only and adaptive), alongside a centralized baseline.
Results on real-world breast cancer imaging data using a pathological complete response prediction model shows that all multi-center schemes outperform centralized monitoring, with F1-score improvements of 1.4–10.3% in drift detection. In the absence of site-specific references, the adaptive scheme performs best, with F1-scores of 74.3% for drift detection and 83.7% for drift severity classification. These findings suggest that adaptive, site-aware agent-based drift monitoring can enhance reliability of multisite clinical decision support systems.

**Keywords:** Clinical decision support system, Drift detection, Intelligent agents.

## 1 Introduction

Translating machine learning predictive models into fully operational clinical decision support systems (CDSS) remains a technically demanding, and resource-intensive endeavor [1]. Even after rigorous validation on independent and external cohorts, these models may still experience a subtle yet impactful drop in performance—referred to as model drift— when deployed in real-world hospitals and medical imaging centers.

A major contributor to this phenomenon is the inherently dynamic and heterogeneous nature of clinical environments, where frequent and multifactorial changes in imaging hardware, acquisition protocols and patient demographics are commonly observed. These changes may impact on the performance of the predictive models, particularly when such variations were not well represented during the models' development [2].

A common strategy to monitor model performance degradation is to track output drifts, i.e., changes in the distribution of model outputs over time. Such changes may indicate an underlying dataset shift [3]—a difference in the joint distribution $P(X, Y)$ between training and deployment—but they can also result from other factors, such as model updates, data preprocessing changes, or system errors.

Recent advances in data de-identification, secure communication protocols, and privacy-preserving cloud infrastructures have enabled multisite CDSS—platforms to serve decision support to multiple and independent healthcare institutions [4]. However, drift detection systems typically rely on centralized monitoring schemes that aggregate data globally [5]. This centralization may obscure localized drifts that compromise the model's reliability in specific institutions.

Software agents have recently sparked renewed attention in AI [6], leveraging distributed decision-making and modular task execution. Only a few agent-based systems have been applied to drift monitoring. One example is [7], which employs an ensemble of agents to detect drift in healthcare signals. Nevertheless, continuous, unsupervised multi-site agent-based drift monitoring in medical imaging remains largely unexplored.

In this study, we propose an agent-based drift monitoring framework suitable for multisite CDSS with dedicated drift monitoring agents to each clinical site. This approach aims to enable adaptable and localized detection of drift while also supporting global aggregation of drift signals across all agents. We validated this framework through an in-silico simulation using real-world breast cancer imaging data [13], for output drift detection in predicting pathological complete response (pCR) to neoadjuvant chemotherapy (NAC). Our hypothesis is that the proposed agent-based monitoring framework for multisite CDSS enables accurate, fine-grained, site-specific drift detection and facilitates a unified, time-aware assessment of drift severity across institutions.

## 2  Materials and Methods

This section presents the agent-based drift monitoring framework for multisite CDSS, outlining the conceptual design of the monitoring agents, and describing an in-silico simulation for output drift detection.

### 2.1  Drift Monitoring Agent

A drift monitoring (DRM) agent is a dedicated monitoring process responsible for overseeing the integrity of a predictive model being used by an individual clinical center. Conceptually, it is a lightweight, continuously operating software component, that ingests model input and/or output data (in real time or batch regime) and applies a drift detection method to assess the presence of change over time. A modular design is proposed for the DRM agents, in which the drift detection logic is decoupled from the core monitoring process. This separation of concerns enables flexibility in configuring the monitoring system. Inspired by the MAPE-K (Monitor, Analyze, Plan, and Execute) control loop of self-managing systems [8], we propose four distinct architectural steps (i.e., initialization, perception, reason and act) for a DRM agent (Figure 1).

**Fig. 1.** Main tasks involved on a DRM agent A$_{<center,\ model>}$.

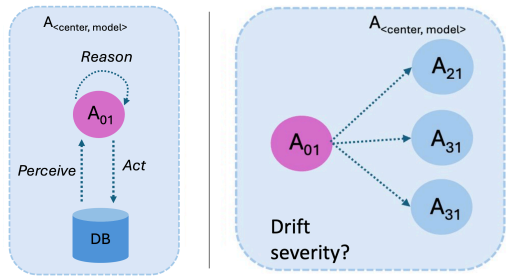

**Initialization.** The initialization (or registration) of a DRM agent involves specifying the predictive model, the clinical center, and the selected drift detection method with its configuration parameters. Additionally, each agent is assigned with a monitoring regime (i.e., on-line, batch) together with a window size and an evaluation frequency, enabling personalization to the data throughput characteristics of the clinical site.

**Perception.** Depending on the monitoring regime, the agent accumulates incoming observations until the predefined monitoring window is reached (batch mode) or processes each observation sequentially upon arrival (online mode).

**Reasoning.** This function defines the core action of the agent, that is the detecting drift within the current monitoring window. This task is determined by the drift method with which the agent was initialized. The expected output from this step encompasses a drift score (usually a p-value), together with a binary drift value (obtained via thresholding) that signals whether drift is present.

- **Drift severity detection.** Multiple DRM agents can concurrently monitor the same predictive model but different sites. We propose a collaborative score based on consensus among DRM agents. Specifically, for a given predictive model $M$ and clinical center $i$, the drift severity score at time t is defined as:

$$S_M(t) = \frac{1}{N_M} \sum_{i=1}^{N_M} d_i(t) \quad (1)$$

where $N_M$ is the number of agents monitoring model $M$, and $d_i(t) \in \{0,1\}$ indicates whether agent $i$ has detected drift at time $t$.

**Action.** Upon drift detection, the agent stores relevant results (e.g., predictions, drift scores) in a database (DB) and triggers tailored actions based on the detected drift and severity, such as notifying external stakeholders or interfacing with auxiliary services.

## 2.2 Output Drift Detection Simulation

To illustrate the applicability of the proposed DRM agent framework, we design a controlled in-silico simulation to detect unsupervised output drift in an environment composed of multiple clinical sites interacting with a CDSS embedding a predictive model. In the simulation, each site is assigned an independent DRM agent, which monitors the distribution of the model's probabilistic outputs (i.e., predicted class probabilities) using only site-specific data. This design enables evaluation of site-specific (multi-center) monitoring under controlled drift scenarios.

**Output Drift Detection**
To enable continuous surveillance, DRM agents are configured to monitor model outputs (i.e., SoftMax probabilities) in batch mode with non-overlapping window batches. Each DRM agent is equipped with an unsupervised output drift detection mechanism, the two-sample Kolmogorov–Smirnov (KS) test. This method was selected for its non-parametric nature, sensitivity to general distributional changes, and widespread adoption in detection tasks [9]. Since ideal i.i.d. conditions are rarely met in production—due to limited batch sizes, repeated testing, and potential temporal autocorrelation—we employ a bootstrap-based permutation procedure ($B=1000$) to estimate finite-sample p-values. This approach is a more robust alternative to asymptotic approximations and helps to mitigate the inflated false positive rates under non-ideal sampling conditions.

**Monitoring Schemes**
We propose four multi-center monitoring schemes, each differing in how the drift detection method constructs the reference distribution. *GlobalRef* – The reference distribution is computed from model outputs on a global evaluation set (defined during model development and encompassing all sites). *SiteRef* – Each site's reference distribution is computed exclusively from the subset of the global evaluation set corresponding to that site. *ProdRef* – The reference distribution is derived from predictions generated during the initial batch size of the simulated production phase. *AdaptiveRef* – The reference distribution is initialized from the global evaluation set and progressively updated using a weighted combination of historical evaluation outputs and recent production data (see next Subsection). For comparison, we additionally include a *Centralized* scheme, in which a single DRM agent monitors outputs aggregated across all sites being compared with the global evaluation set.

**Adaptive Reference Updating**

Building upon progressive adaptation drift detection algorithms [10-12], we propose an adaptive monitoring scheme for DRM agents operating within multisite CDSS. This approach leverages a global reference distribution established during model development, which is gradually adapted using empirical data from the target site while controlling for contamination by drifted observations. The proposed method operates as follows:

- **Initialization**: A global reference distribution $P_{global}$ is constructed from the model's output probabilities on a representative evaluation dataset. This distribution serves as the initial reference for all data centers.
- **Histogram-based Distribution Estimation**: Both the global reference and the empirical data observed at the center are transformed into discrete probability distributions using histograms with a fixed number of bins $K$ over a predefined range (i.e, [0,1]). This discretization facilitates distribution comparison using non-parametric statistical tests.
- **Progressive Blending of References**: As more data becomes available from a specific center, the method constructs an empirical distribution $P_{center}$ from recent model output probabilities. A *blended reference distribution* $P_{ref}$ is then computed as a convex combination of the global and center-specific distributions:

$$P_{\text{ref}} = \lambda \cdot P_{\text{global}} + (1 - \lambda) \cdot P_{\text{center}}, \quad (2)$$

where $\lambda$ controls the relative influence of the global reference, with a decay over time, gradually increasing the weight of the center-specific distribution.

- **Drift Detection via Statistical Testing**: To ensure comparability among schemes, the method also applies the KS test with bootstrap permutation ($B=1000$) to compare the current monitoring batch against the blended reference $P_{ref}$. A statistically significant difference (e.g., p-value below a predefined threshold) is interpreted as potential drift.
- **Controlled Reference Update**: To prevent contamination by drifted data, the blended reference $P_{ref}$ and decay parameter $\lambda$ are updated only when no drift is detected, and the current drift p-value is lower than the previously recorded value. This ensures that the adapted reference evolves only under stable conditions, thereby maintaining robustness to transient or spurious changes in the data distribution.

**Simulation Pipeline**

We implemented a parameterized pipeline (see section 3.2) that configures and executes all five different monitoring schemes through six sequential steps.

- *Data selection:* load input series and reference data for each clinical site. We assume input series are free of pre-existing drift, collected under controlled conditions.
- *Data augmentation:* Bootstrap input series $D$-times to balance sample sizes.
- *Drift injection:* Random injection of drift in the augmented series. Drift values are generated by repeatedly drift duration ($Du$) sampling a uniform distribution defined by the input series' mean and standard deviation scaled by a drift strength ($Sr$).

– *Data sparsity:* Equalize input series length by randomly padding null values.
– *Monitoring setup:* Initialize DRM agents per monitoring scheme and parameters
– *Evaluation:* Execute monitoring schemes. Start the DRM agents, and compute performance metrics (ground truth is known with the location of the injected drift).

## 3  Data and Evaluation

We conducted an in-silico evaluation of the framework for output drift detection in pCR prediction for NAC, using a radiomics-based model built on the MAMA-MIA cohort [13], comprised of retrospectively annotated breast cancer magnetic resonance images. Model, data partitions and radiomic extraction procedures were provided by the Radioval EC project (Id 101057699) [14].

### 3.1  Dataset

The study data consisted of prediction probabilities for the positive class of the binary pCR model, obtained from the evaluation and test partitions used during model development. The evaluation set served as the global reference, while the test set was evenly split into training and testing subsets for monitoring simulations. To emulate a multi-center scenario, both sets were divided into four subsets (Table-1), each corresponding to the one of the four distinct data sources that comprise the Mamma-Mia cohort [13].

**Table 1.** Reference and Test partitions used for the drift detection monitoring simulations

|           | DS-0 | DS-1 | DS-2 | DS-3 | All |
|-----------|------|------|------|------|-----|
| Reference | 39   | 171  | 11   | 14   | 235 |
| Test      | 92   | 128  | 64   | 18   | 300 |

### 3.2  Experiments

We configured 27 pipeline simulations by combining three window sizes, drift strengths, and drift durations across five monitoring schemes (parameterizations summarized in Table-2). Each multi-center scheme employed four DRM agents, each monitoring a distinct subset (DS-X) of the training and testing cohort.

**Table 2.** Parameterization for the different simulations.

| Parameters      |        | Values         | Details                                                  |
|-----------------|--------|----------------|----------------------------------------------------------|
| Augmentation    | (*D*)  | 10%            | Over the size of the input series                        |
| Drift Threshold | (*Th*) | 0.05           | For all drift detection methods                          |
| Drift strength  | (*Sr*) | 20%\|30%\|50%  | Over the probability mean of the augmented input series  |
| Drift duration  | (*Du*) | 20%\|30%\|50%  | Over the size of the augmented input series              |
| Window size     | (*Ws*) | 5%\|10%\|15%   | Over the size of the sparse input series                 |

### 3.3 Metrics

All pipeline simulations were run on training and testing partitions using Monte Carlo sampling (N = 500), enabling estimation of mean and standard deviation for precision, recall, specificity, and F1-score. Each batch processed by a monitoring agent was evaluated as true positive (TP), false positive (FP), true negative (TN), and false negative (FN), for both drift detection and severity classification. For binary drift detection, TP indicated correct detection when drift was present, FP incorrect detection when absent, TN correct identification of no drift, and FN missed detection. For drift severity classification, TP corresponded to correct identification of drift affecting multiple agents in the same batch, FP to overestimation, FN to underestimation, and TN to correct identification of no simultaneous drift.

## 4 Results

This section provides the test results obtained from executing all simulations experiments on the test set. Table-3 summarizes the results for the different monitoring strategies. All multi-center strategies outperformed the *centralized* scheme, with improvements from 1.4% (*ProdRef*) to 10.3% (*SiteRef*) in F1-score. The *SiteRef* obtained the highest performance with 0.774, followed by *AdaptiveRef* with 0.743 of F1-score.

**Table 3.** Drift detection test performance across monitoring schemes. Multi-center results represent the mean (± std) of performance metrics aggregated across all DRM agents.

|             | F1-score (Train) | Precision     | Sensitivity   | Specificity   | F1-score      |
|-------------|------------------|---------------|---------------|---------------|---------------|
| Centralized | 0.746±0.125      | **0.912±0.074** | 0.551±0.179 | **0.927±0.061** | 0.671±0.150 |
| GlobalRef   | 0.711±0.085      | 0.632±0.226   | **0.901±0.171** | 0.652±0.327 | 0.723±0.195   |
| SiteRef     | **0.836±0.065**  | 0.746±0.226   | 0.862±0.219   | 0.864±0.141   | **0.774±0.187** |
| ProdRef     | 0.679±0.244      | 0.756±0.403   | 0.657±0.352   | 0.799±0.387   | 0.685±0.368   |
| AdaptiveRef | 0.761±0.086      | 0.713±0.233   | 0.818±0.227   | 0.793±0.261   | 0.743±0.206   |

Figure-2 shows the average performance of the four drift monitoring agents for each multi-center strategy, and Figure-3 a detailed breakdown by drift and window sizes.

**Fig. 2.** Comparison of multi-center monitoring schemes, stratified by monitoring agents

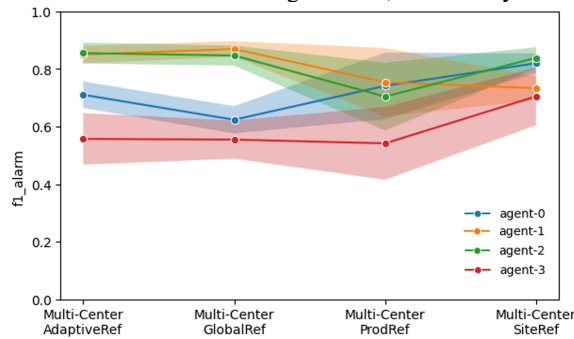

**Fig. 3.** Comparison of multi-center schemes break-down by drift types and window-size conditions

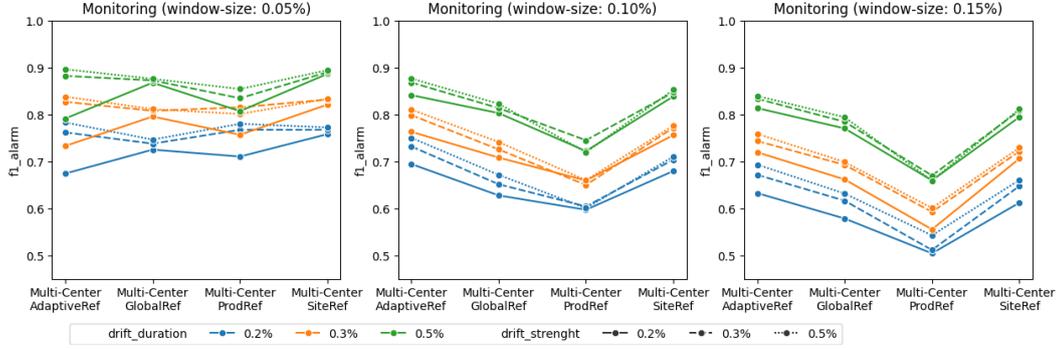

**Table 4.** Drift severity test performance for the different multi-centric strategies. Multi-center results represent the mean (± std) of performance metrics aggregated across all DRM agents.

|  | F1-score (Train) | Precision | Sensitivity | Specificity | F1-score |
|---|---|---|---|---|---|
| GlobalRef | 0.692±0.215 | 0.645±0.242 | **0.982±0.123** | 0.445±0.302 | 0.753±0.209 |
| SiteRef | **0.866±0.197** | **0.812±0.203** | 0.978±0.134 | **0.798±0.199** | **0.876±0.172** |
| ProdRef | 0.746±0.352 | 0.738±0.370 | 0.801±0.338 | 0.727±0.372 | 0.746±0.355 |
| AdaptiveRef | 0.770±0.227 | 0.774±0.24 | 0.956±0.169 | 0.728±0.283 | 0.837±0.207 |

Regarding drift severity detection, Table-4 summarizes the performances of each multi-center monitoring strategy. The *SiteRef* strategy obtained the highest F1-score 87.6%, followed by the *AdaptiveRef* strategy 83.7%. Figure-4 presents qualitative results obtained from the execution of the *AdaptiveRef* strategy.

**Fig. 4.** Qualitative drift detection results. On the left, we observe the data for 4 different centers, predicted and real drifts. On the right side, predicted and real drift severity detection results.

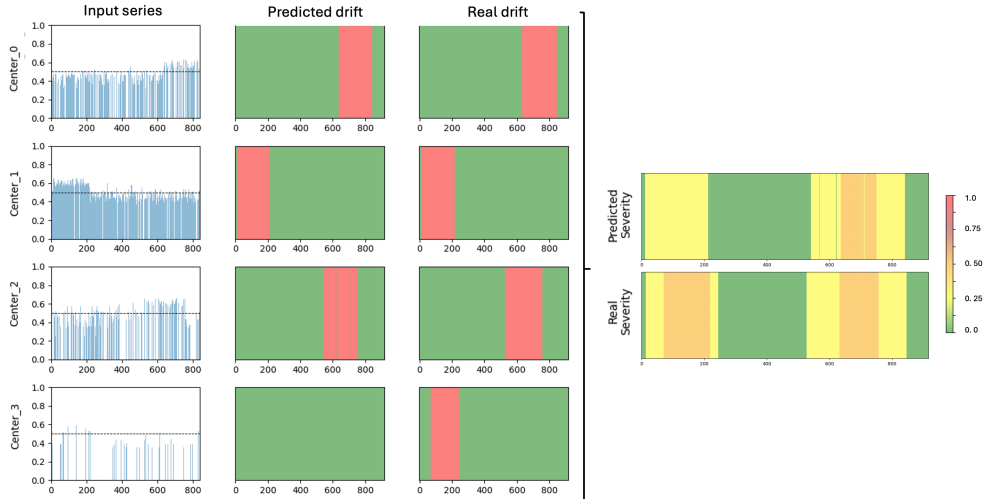

## 5 Discussion

This work introduced an agent-based framework for drift detection in multisite CDSS. Unlike conventional centralized approaches [5], the framework is inherently scalable, as new monitoring agents can be instantiated for additional predictive models or clinical sites, without altering the existing infrastructure. Its modular design also supports maintainability and flexibility as each agent encapsulates a specific drift detection method, facilitating independent updates, method versioning, or replacement across the network. Furthermore, the framework facilitates inter-agent communication that yields collective insights, such as severity scoring from cross-site drift convergence.

We evaluated the framework in-silico for unsupervised output drift detection in pCR prediction for breast cancer patients undergoing NAC. The study compared four multi-center monitoring schemes with a centralized approach, each differing in the origin of the reference data. Across a comprehensive set of simulation experiments, the results (consistent with training) showed that all multi-center schemes outperformed the centralized strategy in terms of F1-score for drift detection. Among the multi-center monitoring schemes, *SiteRef* was the best strategy in F1-score for both drift detection and severity scoring, respectively. This result was expected, as *SiteRef* used site-specific reference distributions, facilitating more accurate drift identification. Among the multi-center schemes without site-specific reference data, an assumption more realistic for production deployment, *AdaptiveRef* outperformed *ProdRef* and *GlobalRef* by 5.9% and 2% in F1-score, respectively, and surpassed both by nearly 9% in severity drift detection. Although various parameter configurations were empirically tried during training, these results (obtained with $K$-bins=100, $\lambda$=1, and decay=0.1) highlight the benefit of integrating production data into the reference distribution via histogram blending, weight decay and conditional updates—a novel combination based on already existing sliding-window adaptive drift detection algorithms [10-12]. Nonetheless, a comprehensive comparison with such algorithms is required to establish its effectiveness.

The performance impact of drift detection within the multi-center schemes is shown in Figure-3. Our results indicate that larger and more pronounced drifts are associated with improved F1-scores for both drift and severity detection. Conversely, smaller drifts tend to yield lower performance metrics. Additionally, we observed a general decline in performance with increasing monitoring window sizes. This trend may be influenced by the abrupt nature of the configured drift, which could hinder the system's adaptability. Further investigation is warranted to validate this hypothesis.

This study has several limitations. First, the simulation of the agent-based framework was conducted in silico, rather than in a real-world deployment, which may limit generalizability. Second, while the designed simulation detected output drift, it did not identify its causes nor model performance degradation. Finally, a more comprehensive fine-tuning and testing is needed, involving a larger number of agents, additional monitoring parameterizations and drift configurations. Future research is envisaged exploring novel collaborative strategies among distinct DRM agents to enhance drift and severity detection as well as improve the interpretability of results.

## 6 Conclusions

We propose an agent-based drift monitoring framework tailored for multisite CDSS, featuring modular, site-specific monitoring agents. Through in-silico evaluation on pCR prediction in breast cancer, we demonstrated that multi-center monitoring consistently surpassed a centralized approach in detecting output drift. Notably, our adaptive scheme, designed for scenarios lacking site-specific reference data, achieved superior drift detection and drift severity scoring—highlighting its practical value for real-world deployments.